\documentclass[11pt]{article}
\usepackage{graphicx}
\usepackage{color}
\usepackage{amsmath}
\usepackage{amssymb}
\usepackage{amscd}
\usepackage{bbm}
\newcommand{\R}{\mathbb{R}}
\newcommand{\inr}[1]{\bigl< #1 \bigr>}

\newcommand{\E}{\mathbb{E}}

\newcommand{\eps}{\varepsilon}

\newcommand{\cN}{{\cal N}}
\newcommand{\cM}{{\cal M}}

\newtheorem{Theorem}{Theorem}[section]
\newtheorem{Lemma}[Theorem]{Lemma}

\newtheorem{Definition}[Theorem]{Definition}

\newtheorem{Remark}[Theorem]{Remark}

\newtheorem{Question}[Theorem]{Question}
\numberwithin{equation}{section}

\def \proof {\noindent {\bf Proof.}\ \ }

\def \endproof
{{\mbox{}\nolinebreak\hfill\rule{2mm}{2mm}\par\medbreak}}
\def\IND{\mathbbm{1}}

\begin{document}
\title{`local' vs. `global' parameters -- breaking the gaussian complexity barrier}
\author{Shahar Mendelson\footnote{Department of
Mathematics, Technion, I.I.T, Haifa 32000, Israel
\newline
{\sf email: shahar@tx.technion.ac.il}
\newline
Supported in part by the Mathematical Sciences Institute,
The Australian National University, Canberra, ACT 2601,
Australia. Additional support was given by the Israel Science Foundation grant 900/10.
 } }

\medskip
\maketitle
\begin{abstract}
We show that if $F$ is a convex class of functions that is $L$-subgaussian, the error rate of learning problems generated by independent noise is equivalent to a fixed point determined by `local' covering estimates of the class, rather than by the gaussian averages. To that end, we establish new sharp upper and lower estimates on the error rate for such problems.
\end{abstract}

\section{Introduction} \label{sec:intro}
The focus of this article is on the question of {\it prediction}. Given a class of functions $F$ defined on a probability space $(\Omega,\mu)$ and an unknown target random variable $Y$, one would like to identify an element of $F$ whose `predictive capabilities' are (almost) the best possible in the class. The notion of `best' is measured via the point-wise cost of predicting $f(x)$ instead of $y$, and the best function in the class is the one that minimizes the average cost.
Here, we will consider the squared loss: the cost of predicting $f(x)$ rather than $y$ is $(f(x)-y)^2$, and if $X$ is distributed according to $\mu$, the goal is to identify
$$
f^*={\rm argmin}_{f \in F} \E(f(X)-Y)^2 = {\rm argmin}_{f \in F} \|f-Y\|_{L_2}^2,
$$
where the expectation is taken with respect to the joint distribution of $X$ and $Y$ on the product space $\Omega \times \R$.

The information at one's disposal is rather limited: a random sample $(X_i,Y_i)_{i=1}^N$, selected according to the $N$-product of the joint distribution of $X$ and $Y$. And, using this data, one must select some (random) $f \in F$.

\begin{Definition} \label{def:basic-def}
Given a sample size $N$ and a class $F$ defined on $(\Omega,\mu)$, a learning procedure is a map $\Psi:(\Omega \times \R)^N \to F$. For a set ${\cal Y}$ of admissible targets, $\Psi$ performs with confidence $1-\delta$ and accuracy ${\cal E}_p$ if for every $Y \in {\cal Y}$, and setting $\tilde{f}=\Psi((X_i,Y_i)_{i=1}^N)$,
$$
\E\left((\tilde{f}-Y)^2|(X_i,Y_i)_{i=1}^N\right) \leq \E(f^*(X)-Y)^2 + {\cal E}_p
$$
with probability at least $1-\delta$ relative to the $N$-product of the joint distribution of $X$ and $Y$.
\end{Definition}
The accuracy (or error) ${\cal E}_p$ is a function of $F$, $N$ and $\delta$, and may depend on some features of the target $Y$ as well, for example, its norm in some $L_q$ space.

\vskip0.4cm

A fundamental problem in Learning Theory is to identify the features of the underlying class $F$ and of the set of admissible targets ${\cal Y}$ that govern ${\cal E}_p$; in particular, the way ${\cal E}_p$ scales with the sample size $N$ (the so-called {\it error rate}). This question has been studied extensively, and we refer the reader to the manuscripts \cite{MR1240719,MR2319879,MR2829871,MR2724359,MR2807761,MR2807761,Men-LWC,Men-LWCG} for more information on its history and on some more recent progress.

\vskip0.3cm

Here, the aim is to obtain matching upper and lower bounds on ${\cal E}_p$ that hold for any reasonable class $F$, at least under some assumptions which we will now outline.

\vskip0.3cm

It is well understood that the ability to predict is quantified by various complexity parameters of the underlying class. Frequently, one encounters parameters that are based on various gaussian and empirical/multiplier processes indexed by `localizations' of $F$ (see, e.g., \cite{LM13}), and any hope of obtaining matching bounds on ${\cal E}_p$ must be based on sharp estimates on these processes. Unfortunately, the analysis of empirical/multiplier processes is, in general, highly nontrivial. Moreover, and unlike gaussian processes, there is no clear path that leads to sharp bounds on empirical processes, and even when upper estimates are available, they are often loose and lead to suboptimal bounds on ${\cal E}_p$.

The one generic example in which a more satisfactory theory of empirical/multiplier processes is known, is when the indexing class is $L$-subgaussian.
\begin{Definition} \label{def:subgaussian-class}
A class $F \subset L_2(\mu)$ is $L$-subgaussian with respect to the measure $\mu$ if for every $p \geq 2$ and every $f,h \in F \cup \{0\}$,
$$
\|f-h\|_{L_p(\mu)} \leq L \sqrt{p}\|f-h\|_{L_2(\mu)},
$$
and if the canonical gaussian process $\{G_f : f \in F\}$ is bounded (see the book \cite{MR1720712} for a detailed survey on gaussian processes).
\end{Definition}
More facts on subgaussian classes may be found in \cite{LT:91,vanderVaartWellner,MR1720712,MR2373017,LM13}. For our purposes, the main feature of subgaussian classes is that the empirical and multiplier processes that govern ${\cal E}_p$ may be bounded from above using properties of the canonical gaussian process indexed by the class, giving one some hope of obtaining sharp estimates. Because of that feature, we will focus in what follows on subgaussian classes.

\vskip0.3cm

Despite their importance, complexity parameters are not the entire story when it comes to ${\cal E}_p$. For example, it is possible to construct a class consisting of just two functions, $\{f_1,f_2\}$, but if the target $Y$ is a $1/\sqrt{N}$-perturbation of the midpoint $(f_1+f_2)/2$, no learning procedure can perform with an error that is better than $c/\sqrt{N}$ having been given a sample of cardinality $N$ (see, e.g., \cite{MR1741038}). Thus, rather than being solely determined by the complexity of the underlying class, there is an additional geometric requirement on $F$ and ${\cal Y}$ which is there to ensure that all the admissible targets in ${\cal Y}$ are located in a favourable position relative of $F$ (see \cite{MR2426759} for more details). One may show that if $F \subset L_2(\mu)$ is compact and convex, any target $Y \in L_2$ is in a favourable position relative to $F$. Therefore, to remove possible geometric obstructions, we will assume that $F \subset L_2(\mu)$ is compact and convex.
\vskip0.3cm

Finally, for a reason that will become clear later, we will not study a general class of admissible targets ${\cal Y}$, but rather consider targets of the form $Y=f(X)+W$ for some $f \in F$ and $W$ that is orthogonal to ${\rm span}(F)$ (e.g., $W \in L_2$ that is a mean-zero random variable and is independent of $X$ is a `legal' choice).
\vskip0.3cm

With all these assumptions in place, let us formulate the question we would like to study:
\begin{Question} \label{qu:main}
Let $F \subset L_2(\mu)$ be a compact, convex class that is $L$-subgaussian with respect to $\mu$. Given targets of the form $Y=f(X)+W$ as above, find matching upper and lower bounds (up to constants) on ${\cal E}_p$.
\end{Question}
Let us recall the following standard definitions.
\begin{Definition}
Let $F \subset L_2(\mu)$. Set
$$
F-h=\{f-h: f \in F\} \ \ {\rm and} \ \ F-F=\{f-h: f,h \in F\}.
$$
Denote by
$$
{\rm star}(F)=\{\lambda f  \ : \ f \in F \ \ 0 \leq \lambda \leq 1\}
$$
the star-shaped hull of $F$ with $0$; $F$ is star-shaped around $0$ if ${\rm star}(F)=F$.

Let $\{G_f : f \in F\}$ be the canonical gaussian process indexed by $F$ and set
$$
\E\|G\|_F = \sup\left\{ \E \sup_{f \in F^\prime} G_f  \ : \ F^\prime \subset F, \ F^\prime \ {\rm is \ finite} \ \right\}.
$$
Finally, let $D$ be the unit ball in $L_2(\mu)$.
\end{Definition}

The best known bounds on ${\cal E}_p$ in the subgaussian context have been established in \cite{LM13} and are based on two fixed points:
\begin{Definition} \label{def:gaussian-parameters}
For $\kappa_1, \kappa_2>0$, set
\begin{equation} \label{eq:app-upper-r-M}
r_M(\kappa_1,f)=\inf \left\{s>0 :  \E\|G\|_{(F-f) \cap sD} \leq \kappa_1 s^2 \sqrt{N} \right\}
\end{equation}
and
\begin{equation} \label{eq:app-upper-r-Q}
r_Q(\kappa_2,f)=\inf \left\{r>0 : \E\|G\|_{(F-f) \cap sD} \leq \kappa_2 s \sqrt{N} \right\}.
\end{equation}
Put
$$
r_M(\kappa_1) = \sup_{f \in F} r_M(\kappa_1,f) \ \ {\rm and} \ \ r_Q(\kappa_2)=\sup_{f \in F} r_Q(\kappa_2,f).
$$
\end{Definition}
In the context of the problem we are interested in, one has the following:
\begin{Theorem} \label{thm:LM-main-upper} \cite{LM13}
For every $L \geq 1$ there exist constants $c_1,c_2,c_3$ and $c_4$ that depend only on  $L$ for which the following holds. Let $F \subset L_2(\mu)$ be a compact, convex, $L$-subgaussian class of functions, set $Y=f_0(X)+W$ and assume that for every $p \geq 2$, $\|W\|_{L_p} \leq L\sqrt{p}\|W\|_{L_2}$. There is a learning procedure (empirical risk minimization performed in $F$) for which, if
$$
r \geq 2\max\left\{r_M(c_0/\|W\|_{L_2}),r_Q(c_1)\right\}.
$$
then with probability at least
$$
1-2\exp\left(-c_2N \min\left\{1,r^2/\|W\|_{L_2}^2\right\}\right),
$$
the error of the procedure is at most ${\cal E}_p \leq r^2$.
\end{Theorem}

The lower bound that complements Theorem \ref{thm:LM-main-upper} uses `local' analogs of $r_M$ and $r_Q$ that are based on the notion of packing numbers.

\begin{Definition} \label{def:packing-numbers}
Let $E$ be a normed space and set $B$ to be its unit ball. Let ${\cal M}(A,rB)$ be the cardinality of a maximal $r$-separated subset of $A$ with respect to the given norm, that is, the cardinality of the largest subset $(a_i)_{i=1}^m \subset A$ for which $\|a_i-a_j\| \geq r$ for every $i \not= j$.
\end{Definition}

\begin{Definition} \label{def:sudakov-complexity-thm}
For $\eta_1,\eta_2>0$ set
$$
\gamma_M(\eta_1,f) = \inf \left\{s>0 : \log \cM\left((F-f) \cap 4sD, (s/2)D\right) \leq \eta_1^2 s^2 N \right\}.
$$
and
$$
\gamma_Q(\eta_2,f)= \inf \left\{s>0 : \log \cM\left((F-f) \cap 4sD, (s/2) D\right) \leq \eta_2^2 N \right\}.
$$
Put
$$
\gamma_M(\eta_1)=\sup_{f \in F} \gamma_M(\eta_1,f), \ \ {\rm and} \ \ \gamma_Q(\eta_2)=\sup_{f \in F} \gamma_Q(\eta_2,f).
$$
\end{Definition}

\begin{Theorem} \label{thm:LM-main-lower} \cite{LM13}
There exist absolute constants $c_1$ and $c_2$ for which the following holds. Let $F$ be a class of functions, set $W$ be a centred normal random variable and for every $f \in F$ put $Y^f=f(X)+W$. If $\Psi$ is a learning procedure that performs for every target $Y^f$ with confidence at least $3/4$, then there is some $Y^f$ for which ${\cal E}_p \geq c_1 \gamma_M^2(c_2/\|W\|_{L_2})$.
\end{Theorem}

\begin{Remark}
One should note that a lower bound that is based on $\gamma_Q$ was not known.
\end{Remark}

The connection between the two types of parameters is Sudakov's inequality (see, e.g. \cite{LT:91}):  there is an absolute constant $c$ for which, for every $H \subset L_2(\mu)$,
$$
c\sup_{\eps>0} \eps \log^{1/2} \cM(H,\eps D) \leq \E\|G\|_H.
$$
To see the connection, assume that for every $f \in F$, $\E\|G\|_{(F-f) \cap 4rD} \leq \kappa_1 (4r)^2 \sqrt{N}$, which means that $r_M(\kappa_1) \leq 4r$. Applying Sudakov's inequality to $H=(F-f) \cap 4rD$ and for the choice of $\eps=r/2$,
$$
c(r/2)\log^{1/2}\cM\left((F-f)\cap 4rD,rD\right) \leq \E\|G\|_{(F-f)\cap 4rD} \leq 16\kappa_1 r^2 \sqrt{N};
$$
hence, $\gamma_M(c_1 \kappa_1) \leq r$. A similar observation is true for $r_Q$ and $\gamma_Q$, which shows that $\gamma_M$ and $\gamma_Q$ are intrinsically smaller than $r_Q$ and $r_M$ respectively, for the right choice of constants.

\vskip0.3cm

The starting point of this article is fact that the gap between these upper and lower estimates on ${\cal E}_p$ is more than a mere technicality.

The core issue is that the parameters $r_M$ and $r_Q$ are `global' in nature, whereas $\gamma_M$ and $\gamma_Q$ are `local'. Indeed, although $(F-f) \cap rD$ is a localized set, $\E\|G\|_{(F-f) \cap rD}$ is not determined solely by the effects of a `level' that is proportional $r$. For example, it is straightforward to construct examples in which  $\E\|G\|_{(F-f) \cap rD} \geq c r\sqrt{N}$ because of a very large, $\rho$-separated subset of $(F-f) \cap rD$, for $\rho$ that is much smaller than $r$. Thus, even if $r_M$ or $r_Q$ are of order $r$, this need not be `exhibited' by $(F-f) \cap rD$ at a scale that is proportional to $r$.  In contrast, $\gamma_M$ and $\gamma_Q$ are `local': the degree of separation is proportional to the diameter of the separated set, and the fixed point indicates that $(F-f) \cap rD$ is truly `rich' at a scale that is proportional to $r$.

As noted in \cite{LM13}, the upper and lower estimates coincide when the `local' and `global' parameters are equivalent, but that is not a typical situation -- in the generic case, there is a gap between the two. An example of that fact will be presented in Section \ref{sec:remarks}.

Given that there is a gap between the two sets of parameters, one must face the obvious question: which of the two captures ${\cal E}_p$? Is it the `global' pair, $r_Q$ and $r_M$, or the `local' one of $\gamma_Q$ and $\gamma_M$?

Our main result is that the `local' parameters are the right answer -- at least in the setup outlined above. To that end, we shall improve the upper bound in Theorem \ref{thm:LM-main-upper} and add the missing component in Theorem \ref{thm:LM-main-lower}.

\begin{Theorem} \label{thm:main-upper}
For every $L>1$ and $q>2$ there are constants $c_0,...,c_5$ that depend only of $q$ and $L$ for which the following holds.
Let $F \subset L_2(\mu)$ be a compact, convex, $L$-subgaussian class of functions with respect to $\mu$. There is a learning procedure $\Psi:(\Omega \times \R)^N \to F$, for which, if $Y=f(X)+W$ for $f \in F$ and $W \in L_q$ that is orthogonal to ${\rm span}(F)$, then with probability at least
$$
1-2\exp\left(-c_0N\min\{1,\gamma_M^2(c_1/\|W\|_{L_q})\}\right)-c_2\frac{\log^q N}{N^{(q/2)-1}},
$$
$$
{\cal E}_p \leq c_3 \max\left\{\gamma_M^2\left(\frac{c_1}{\|W\|_{L_q}}\right), \gamma_Q^2(c_4)\right\} + r_Q^2(c_4)\exp\left(-c_5\exp(N)\right)
$$
\end{Theorem}

The term $r_Q^2(c_4)\exp(-c_5\exp(N))$ is almost certainly an artifact of the proof, but in any case, it is significantly smaller than the dominating term in any reasonable example.

\vskip0.3cm

To complement Theorem \ref{thm:main-upper} we obtain the following lower bound.

\begin{Theorem} \label{thm:main-lower}
There exist absolute constants $c_0$ and $c_1$ for which the following holds. Let $F \subset L_2(\mu)$ be a convex, centrally-symmetric class of functions and let $\Psi$ be any learning procedure that performs with confidence $7/8$ for any target of the form $Y=f(X)+W$ for some $f \in F$ and $W \in L_2$ that is orthogonal to ${\rm span}(F)$.
\begin{description}
\item{$\bullet$} For any $W \in L_2$ that is orthogonal to ${\rm span}(F)$, there is some $f \in F$, for which, for $Y=f(X)+W$,
$$
{\cal E}_p \geq c_0 \gamma_Q^2(c_1).
$$

\item{$\bullet$} If $W$ is a centred, normal random variable that is independent of $X$, there is some $f \in F$ for which, for $Y=f(X)+W$,
$$
{\cal E}_p \geq c_0 \gamma_M^2\left(\frac{c_1}{\|W\|_{L_2}}\right).
$$
\end{description}
\end{Theorem}

An outcome of Theorem \ref{thm:main-upper} and Theorem \ref{thm:main-lower} is that if $W$ is a centred gaussian random variable that is independent of $X$, then for any convex, centrally-symmetric, $L$-subgaussian class $F$, the upper and lower estimates match (up to the parasitic and negligible term $r_Q^2(c_4)\exp(-c_5\exp(N))$ in the upper bound): when considering targets of the form $Y=f(X)+W$ for $f \in F$,
$$
{\cal E}_p \sim \max\left\{\gamma_Q^2(c_1),\gamma_M^2(c_2/\|W\|_{L_2})\right\}.
$$
\vskip0.3cm

The second part of Theorem \ref{thm:main-lower} follows from Theorem \ref{thm:LM-main-lower}. We have chosen to present a new proof of that fact -- a proof we believe is both instructive and less restrictive than existing proofs. The first part of Theorem \ref{thm:main-lower} is, to the best of our knowledge, new.

\vskip0.3cm

Let us mention that if $F$ happens to be convex and centrally symmetric (i.e. if $f \in F$ then $-f \in F$), what is essentially the `richest' shift of $F$ is the $0$-shift. Indeed, since $F-F=2F$, it is evident that for every $f \in F$
$$
(F-f) \cap 4rD \subset (F-F) \cap 4rD =2(F \cap 2rD).
$$
This makes one's life much simpler when studying lower bounds, as it gives an obvious choice of where to look. Indeed, the `richest' part of $F$ is the hardest part for a learning procedure to deal with -- and that part is a neighbourhood of $0$.

\subsection{The idea of the proof of the upper bound}
The proof of the upper bound is based on the following decomposition of the squared excess loss: let $Y$ be the unknown target and set $f^*={\rm argmin}_{f \in F} \|f-Y\|_{L_2}$. For every $f \in F$, let $\ell_f(X,Y)=(f(X)-Y)^2$ and set
\begin{align} \label{eq:quad-decomp}
{\cal L}_f^F(X,Y)= & (\ell_f - \ell_{f^*})(X,Y)=(f(X)-Y)^2-(f^*(X)-Y)^2 \nonumber
\\
= & 2(f^*(X)-Y)(f-f^*)(X)+(f-f^*)^2(X).
\end{align}
Let $P_N h = \frac{1}{N}\sum_{i=1}^N h(X_i,Y_i)$ and set
$$
\hat{f} = {\rm argmin}_{f \in F} P_N \ell_f= {\rm argmin}_{f \in F} P_N {\cal L}_f^F
$$
to be the empirical minimizer in $F$. The learning procedure that assigns to every sample $(X_i,Y_i)_{i=1}^N$ the empirical minimizer in $F$ is called {\it Empirical Risk Minimization} (ERM).

Clearly, ${\cal L}_{f^*}^F=0$, and thus, for every sample $(X_i,Y_i)_{i=1}^N$,
\begin{equation*}
P_N {\cal L}_{\hat{f}}^F \leq 0,
\end{equation*}
implying that members of the random set $\{f \in F : P_N {\cal L}_f^F > 0\}$ cannot be empirical minimizers. One way of identifying that set is via the decomposition \eqref{eq:quad-decomp}: assume that $(X_i,Y_i)_{i=1}^N$ is a sample for which, if $\|f-f^*\| \geq r$, one has
\begin{equation} \label{eq:quad-intro}
\frac{1}{N} \sum_{i=1}^N (f-f^*)^2 \geq \kappa \|f-f^*\|_{L_2}^2,
\end{equation}
and
\begin{equation} \label{eq:multi-intro}
\left|\frac{1}{N}\sum_{i=1}^N (f^*(X_i)-Y_i)(f-f^*)(X_i) - \E(f^*(X)-Y)(f-f^*)(X) \right| \leq \frac{\kappa}{4} \|f-f^*\|_{L_2}^2.
\end{equation}
Since $F$ is compact and convex, by properties of the metric projection onto a closed convex set in an inner product space,
\begin{equation} \label{eq:conv-type-cond}
\E(f^*(X)-Y)(f-f^*)(X) \geq 0
\end{equation}
for every $f \in F$. Therefore, setting $\xi=f^*(X)-Y$ and $\xi_i=f^*(X_i)-Y_i$,
\begin{align*}
P_N {\cal L}_f^F \geq & \frac{1}{N} \sum_{i=1}^N (f-f^*)^2 -2\left|\frac{1}{N}\sum_{i=1}^N \xi_i(f-f^*)(X_i) - \E\xi(f-f^*)(X) \right|
\\
+& \E \xi(f-f^*)(X) \geq  \kappa-2(\kappa/4) >0
\end{align*}
for every $f \in F$ that satisfies $\|f-f^*\|_{L_2} \geq r$. Thus, if \eqref{eq:quad-intro} and \eqref{eq:multi-intro} hold for the sample $(X_i,Y_i)_{i=1}^N$, then
$$
\left\{f \in F : \|f-f^*\|_{L_2} \geq r \right\} \subset \left\{f \in F : P_N{\cal L}_f^F >0\right\}
$$
implying that $\|\hat{f}-f^*\|_{L_2} < r$.

This argument has been used in \cite{Men-LWC} and was then extended in \cite{Men-LWCG}, showing that
$$
\E ({\cal L}_{\hat{f}}^F|(X_i,Y_i)_{i=1}^N) \leq r^2
$$
-- which is the type of result one is looking for.
\vskip0.3cm
This method of proof leads to the complexity parameters $r_Q$ and $r_M$: the former controls the quadratic component \eqref{eq:quad-intro} and the latter the multiplier component \eqref{eq:multi-intro}. The `global' nature of $r_Q$ and $r_M$, i.e., the fact that the two depend on the gaussian oscillation $\E\|G\|_{(F-f) \cap rD}$ cannot be helped: the oscillations of the quadratic and multiplier processes are highly affected by the `richness' of $F$ around $f^*$ at every `level'.
\vskip0.3cm
A rather obvious idea for improving the upper estimate is `erasing' all the fine structure of $F$, for example, by replacing $F$ with an appropriate separated subset. The difficultly in such an approach is that the geometry of a separated set is problematic, and \eqref{eq:conv-type-cond} will no longer be true for an arbitrary target $Y$. This is why we only consider targets of the form $f(X)+W$ for $f \in F$ and $W$ that is orthogonal to ${\rm span}(F)$. For such targets, a version of \eqref{eq:conv-type-cond} happens to be true even if $F$ is replaced by a separated set.

\vskip0.4cm

The path we will take in proving the upper bound is as follows:
\begin{description}
\item{$\bullet$} Choose a `correct' level $r$ using the parameters $\gamma_M$ and $\gamma_Q$ for well-chosen constants $\eta_1$ and $\eta_2$ that depend only on $q$ and $L$.
\item{$\bullet$} Replace $F$ by $V$, a maximal $r$-separated subset of $F$ with respect to the $L_2(\mu)$ norm, and study ERM in $V$. To that end, set $v_0={\rm argmin}_{v \in V} \|v-Y\|_{L_2}$ and observe that by the orthogonality of $W$ to ${\rm span}(F)$, for every $v \in V$,
    $$
    \left|\E (v_0(X)-Y)(v-v_0)(X)\right| = \left|\E (v_0-f^*)(v-v_0)(X)\right| \leq r\|v-v_0\|_{L_2}.
    $$
Therefore, the empirical excess loss relative to $V$ satisfies
\begin{align*}
P_N {\cal L}_v^V \geq & \frac{1}{N}\sum_{i=1}^N (v-v_0)^2(X_i)
\\
- & 2\left|\frac{1}{N}\sum_{i=1}^N (v_0(X_i)-Y_i)(v-v_0)(X_i) - \E (v_0(X)-Y)(v-v_0)(X)\right|
\\
- & 2r\|v-v_0\|_{L_2}.
\end{align*}
\item{$\bullet$} Next, one may study the corresponding quadratic and multiplier processes indexed by localizations of $V$ and show that with high probability, if $\|v-v_0\|_{L_2} \geq c_1r$ then $P_N {\cal L}_v^V >0$. Thus, ERM performed in $V$ produces $\hat{v}$ for which $\|\hat{v}-v_0\|_{L_2} \leq c_1r$.

\item{$\bullet$} It is possible to show that on the same event, $\|\hat{v}-f^*\|_{L_2} \leq c_2r$. And, using the orthogonality of $W$ to ${\rm span}(F)$ once again, $\E({\cal L}^F_{\hat{v}}|(X_i,Y_i)_{i=1}^N) \leq c_3r^2$, as required.
\end{description}

\section{Preliminaries} \label{sec:pre}

Let us begin with some natation. Throughout, absolute constants are denoted by $c,c_1,...$ etc. Their value may change from line to line. $c(\alpha)$ is a constant that depends only on the parameter $\alpha$. We use $\kappa_1,\kappa_2,\eta_1,\eta_2$ etc. to denote fixed constants whose value remains unchanged throughout the article.

In what follows, we will, at times, abuse notation and not specify the probability space on which each random variable is defined. For example, $\|f-Y\|_{L_2}^2=\E(f(X)-Y)^2$ and integration is with respect to the joint distribution of $X$ and $Y$, while $\|f-f_0\|_{L_2}^2=\E(f-f_0)^2(X)$, in which case integration is with respect to $\mu$.

Next, let us turn to the notions of {\it cover} and {\it covering numbers}.
\begin{Definition} \label{def:covering-numbers}
Let $B$ be a unit ball of a norm. Set $\cN(A,B)$ to be the minimal number of centres $a_1,...,a_n \in A$ for which $A \subset \bigcup_{i=1}^n (a_i+B)$. $(a_i)_{i=1}^n$ is called a cover of $A$ with respect to $B$. An $r$-cover is a cover with respect to the set $rB$.
\end{Definition}
It is standard to verify that if $a_1,...,a_m$ is a maximal separated subset with respect to $B$ then it is also a cover with respect to $B$. Indeed, the maximality of the separated set implies that every point $a \in A$ has some $a_i$ for which $\|a_i-a\| \leq 1$, i.e, $a \in a_i+B$. Therefore, $\cN(A,B) \leq \cM(A,B)$. In the reverse direction, if $a_1,...,a_n$ is a cover with respect to $B$, then each one of the balls $a_i+B$ contains at most one point in any $2$-separated set. Thus, $\cM(A,2B) \leq \cN(A,B)$.

\vskip0.3cm

The following lemma is straightforward but it plays a crucial part in what follows.
\begin{Lemma} \label{lemma:covering}
Let $T \subset W \subset L_2(\mu)$. For $s>r>0$, set
$$
\phi(s,r)=\sup_{w \in W} \cN(T \cap (w+sD),rD).
$$
Then
\begin{description}
\item{1.} $\phi(s,r) \leq \phi(s,s/2) \cdot \phi(s/2,r)$.
\item{2.} If $T$ and $W$ are star-shaped around 0 then
$$
\log \phi(s,r) \leq c_0\log(2s/r) \cdot \log \phi(4r,r)
$$
for a suitable absolute constant $c_0$.
\end{description}
\end{Lemma}

\proof Fix $w \in W$ and let $t_1,...,t_N \in T \cap (w + sD)$ be centres of a minimal $s/2$-cover of that set. For every $1 \leq i \leq N$,
$$
T \cap (w+sD) \cap (t_i + (s/2)D) \subset T \cap (t_i + (s/2)D),
$$
and $\cN(T \cap (t_i + (s/2)D),rD) \leq \phi(s/2,r)$, because $t_i \in T \subset W$. Therefore,
$$
\sup_{w \in W} \cN(T \cap (w+sD),rD) \leq \sup_{w \in W} \cN(T \cap (w+sD), (s/2)D) \cdot \phi(s/2,r).
$$

Turning to the second part of the claim, assume that $T$ and $W$ are star-shaped around $0$. Let $w \in W$, set $t_1,...,t_m$ to be a maximal $s/2$-separated subset of $T \cap (w +sD)$ with respect to the $L_2(\mu)$ norm and put $y_i = (r/s)t_i$. Since $T$ is star-shaped around $0$, $y_i \in T$ and $(y_i)_{i=1}^m$ is an $r/2$-separated subset of $(r/s)w+rD$. For the same reason, $(r/s)w \in W$, and
$$
\cM(T \cap (w+sD),rD) \leq \sup_{v \in W} \cM(T \cap (v+rD),(r/2)D).
$$
Using the standard connection between packing numbers and covering numbers and taking the supremum over $w$,
\begin{align*}
\phi(s,s/2) = & \sup_{w \in W} \cN(T \cap (w+sD),(s/2)D) \leq \sup_{w \in W} \cM(T \cap (w+sD),(s/2) D)
\\
\leq & \sup_{w \in W} \cM(T \cap (w+2rD),rD).
\end{align*}
Iterating the first part of the lemma,
\begin{align*}
\log \phi(s,r) \leq & \log_2(2s/r) \cdot \sup_{w \in W} \log \cM(T \cap (w+4rD),2rD)
\\
\leq &\log_2(2s/r) \cdot \sup_{w \in W} \log \cN(T \cap (w+4rD),rD)
\\
\leq & \log_2(2s/r) \cdot \log \phi(4r,r),
\end{align*}
as claimed.
\endproof

Before we turn to the proof of the upper bound, let us revisit the complexity parameters in question. Since $F$ is a convex class, $F-f$ is star-shaped around $0$; hence, if $s>r$
$$
{\cal M}\left((F-f) \cap 4sD, (s/2)D\right) \leq {\cal M}\left((F-f) \cap 4(r/2)D, rD\right).
$$
In particular, if $\gamma_M(\eta_1,f) <r$ then
$$
\log {\cal M}\left((F-f) \cap 4sD,(s/2)D\right) \leq \eta_1^2 N r^2 \leq \eta_1^2 N s^2,
$$
implying that $\gamma_M(\eta_1,f) < s$ as well.

This simple argument shows that if $r < \gamma_M(\eta_1,f)$ then
$$
\log {\cal M}\left((F-f) \cap 4rD,(r/2)D\right) \geq \eta_1^2 N r^2,
$$
while if $r>\gamma_M(\eta_1,f)$, the reverse inequality holds.

A similar assertion holds for $\gamma_Q$, $r_M$ and $r_Q$; the rather standard proof of these facts, which is almost identical to the argument used above, is omitted.

\section{The upper bound} \label{sec:upper-bound}
Let $F \subset L_2(\mu)$ be a compact, convex class of functions. Fix $r>0$ that will be named later and let $V$ to be a maximal $r$-separated subset of $F$. Note that for every $v_0 \in V$, $F_{v_0}=F-v_0$ is star-shaped around $0$, and ${\rm star}(V-v_0) \subset F-v_0$. Using the notation of Lemma \ref{lemma:covering}, let $T=W=F_{v_0}$, and for $s>2r>0$,
\begin{align*}
& \log \cN\left( \left({\rm star(V-v_0)}\right) \cap sD, rD\right) \leq   \log \cN\left( F_{v_0} \cap sD, rD\right)
\\
\leq & \sup_{x \in F} \log \cN\left( F_{v_0} \cap (x-v_0+sD), rD\right)
\\
\leq & c_0\log (s/r) \sup_{x \in F} \log \cN( F_{v_0} \cap (x-v_0+4rD),rD)
\\
= & c_0\log (s/r) \sup_{x \in F} \log \cN( F \cap (x+4rD),rD).
\end{align*}
Also, observe that $F \cap (x+4rD) \subset \left((F-x) \cap 4rD\right)+x$, implying that
\begin{equation} \label{eq:useful-covering-star}
\log \cN\left( \left({\rm star(V-v_0)}\right) \cap sD, rD\right) \leq c_0\log (s/r) \cdot \sup_{x \in F}\log \cN\left( (F-x) \cap 4rD,rD\right).
\end{equation}
Moreover, the same estimate holds for $(V-v_0) \cap sD$, and since $V-v_0$ is $r$-separated,
\begin{align} \label{eq:usefulcover-nostar}
\log |(V-v_0) \cap sD| = & \log {\cal M}\left((V-v_0) \cap sD,rD\right) \leq \log {\cal N}\left((V-v_0) \cap sD,(r/2)D\right) \nonumber
\\
\leq & \log {\cal N}\left(F_{v_0} \cap sD,(r/2)D\right) \nonumber
\\
\leq & c_0\log (s/r) \cdot \sup_{x \in F}\log \cN\left( (F-x) \cap 4rD,(r/2)D\right) \nonumber
\\
\leq & c_0\log (s/r) \cdot \sup_{x \in F}\log \cM\left( (F-x) \cap 4rD,(r/2)D\right)
\end{align}

With that in mind, fix constants $\eta_1,\eta_2, \kappa_2$ and $\kappa_3$ that will be specified later, and for that choice of constants, let $r>0$ for which
\begin{equation} \label{eq:condition-on-r}
\sup_{x \in F}\log \cM \left((F-x) \cap 4rD, (r/2)D\right) \leq \max \left\{\eta_1^2 Nr^2,\eta_2^2 N\right\},
\end{equation}
and
$$
r \geq r_Q(\kappa_2)\exp(-\kappa_3 \exp(N));
$$
that is,
$$
r \geq \max\left\{\gamma_M(\eta_1),\gamma_Q(\eta_2),r_Q(\kappa_2)\exp(-\kappa_3\exp(N))\right\}.
$$

Let $V$ be a maximal $r$-separated subset of $F$ with respect to the $L_2(\mu)$ norm.
Following the path outlined earlier, the idea is to study ERM in $V$, given the data $(X_i,Y_i)_{i=1}^N$ for $Y=f_0(X)+W$. To that end, one must control the multiplier and quadratic components in the decomposition of the squared loss relative to $V$: if $v_0={\rm argmin}_{v \in V} \|v(X)-Y\|_{L_2}$,
\begin{align*}
{\cal L}_v^V(X,Y)= & (v(X)-Y)^2 - (v_0(X)-Y)^2
\\
= & 2(v_0(X)-Y)(v-v_0)(X) + (v-v_0)^2(X).
\end{align*}

Let us begin with the multiplier component:

\begin{Lemma} \label{lemma:local-multiplier}
Fix $0<\theta<1$, $L>1$ and $q>2$. There exist constants $c_0$, $c_1$ and $c_2$ that depend only on $L$ and $q$ and for which the following holds. Let $F$ be a convex, $L$-subgaussian class, set $\xi \in L_q$ for some $q>2$ and put $\eta_1=c_0 \theta /\|\xi\|_{L_q}$. Then, for every $v_0 \in V$, with probability at least
$$
1-c_1\frac{\log^q N}{N^{((q/2)-1)}} - 2\exp(-c_2\eta_1^2r^2N),
$$
$$
\sup_{\{v \in V : \|v-v_0\|_{L_2} \geq  2r\}} \left|\frac{1}{N}\sum_{i=1}^N \xi_i \frac{v-v_0}{\|v-v_0\|_{L_2}^2}(X_i) - \E \xi\frac{v-v_0}{\|v-v_0\|_{L_2}^2}\right| \leq \theta. $$
\end{Lemma}

The proof of Lemma \ref{lemma:local-multiplier} is based on the following fact from \cite{Men-Bern}.

\begin{Theorem} \label{thm:multiplier}
For $L>1$ and $q>2$ there exist constants $c_0$, $c_1$ and $c_2$ that depend only on $L$ and $q$ for which the following holds. Let $\xi \in L_q$, set $H$ to be an $L$-subgaussian class and denote by $d_H=\sup_{h \in H} \|h\|_{L_2}$. For $w,u \geq 8$, with probability at least
$$
1-c_0w^{-q} N^{-((q/2)-1)}\log^{q} N-2\exp\left(-c_1u^2 \left(\frac{\E\|G\|_H}{Ld_H}\right)^2\right),
$$
$$
\sup_{h \in H} \left|\frac{1}{N} \sum_{i=1}^N \xi_i h(X_i) - \E \xi h \right| \leq c_2Lwu\|\xi\|_{L_q}\frac{\E\|G\|_{H}}{\sqrt{N}}.
$$
\end{Theorem}

\noindent{\bf Proof of Lemma \ref{lemma:local-multiplier}.}
The proof consists of two parts: first, controlling the process indexed by $\{f \in F:\|f-v_0\|_{L_2} \geq s\}$ where $s=(3/2)r_M(\eta_1,v_0)$, and then treating the process indexed by $\{v \in V:r \leq \|v-v_0\|_{L_2} \leq s\}$. Clearly, without loss of generality one may assume that $r \leq r_M(\eta_1,v_0)$.

By the regularity of $r_M$ and since $s>r_M(\eta_1,v_0)$,
$$
\E\|G\|_{(F-v_0) \cap sD} \leq \eta_1 \sqrt{N} s^2.
$$
Moreover, $(F-v_0) \cap (s/4) D \subset (F-v_0)\cap sD$, and since $s/4 \leq r_M(\eta_1,v_0)$, the regularity of $r_M$ implies that
$$
\E\|G\|_{(F-v_0) \cap sD} \geq \eta_1 \sqrt{N} s^2/16.
$$

Therefore, applying Theorem \ref{thm:multiplier} to the set $H=(F-v_0) \cap sD$, there are constants $c_1$, $c_2$ and $c_3$ that depend only on $q$ and $L$ for which, with probability at least
$$
1-c_1N^{-((q/2)-1)}\log^{q} N-2\exp\left(-c_2\eta_1^2 s^2 N  \right),
$$
if $ f \in F$ and $\|f-v_0\|_{L_2} \leq s$,
$$
\left|\frac{1}{N} \sum_{i=1}^N \xi_i (f-v_0)(X_i) - \E \xi(f-v_0)\right| \leq c_3L\|\xi\|_{L_q}\eta_1s^2 =(*).
$$
Clearly, $(*) \leq \theta s^2$ if $\eta_1 \leq \theta/c_3L\|\xi\|_{L_q}$, and for such a choice, if $\|f-v_0\|_{L_2}=s$ then
\begin{equation} \label{eq:multiplier-in-proof-1}
\left|\frac{1}{N} \sum_{i=1}^N \xi_i (f-v_0)(X_i) - \E \xi(f-v_0)\right| \leq \theta\|f-v_0\|_{L_2}^2;
\end{equation}
since $F-v_0$ is star-shaped around $0$, \eqref{eq:multiplier-in-proof-1} holds on the same event for every $f \in F$ for which $\|f-v_0\| \geq s$.

Next, one has to control the process indexed by $\{v \in V: r \leq \|v-v_0\|_{L_2} < s\}$. Set $j_0=\lceil s/r \rceil$, fix $s_j=2^j r$ for $0 \leq j \leq j_0$ and let $V_j={\rm star}((V-v_0)\cap s_jD)$. By Theorem \ref{thm:multiplier}, on an event ${\cal A}_j$, for every $h \in V_j$,
$$
\left|\frac{1}{N}\sum_{i=1}^N \xi_i  h(X_i) - \E \xi h \right| \leq c_4(L,q) w_ju_j\|\xi\|_{L_q}\frac{\E\|G\|_{V_j}}{\sqrt{N}}=(**)_j.
$$
The aim it to ensure that $(**)_j \leq \theta s_j^2/4$ and that ${\cal A}_j$ is of high enough probability. Indeed, on ${\cal A}_j$, if $v \in V$ and $s_j/2 \leq \|v-v_0\|_{L_2} \leq s_j$,
$$
\left|\frac{1}{N}\sum_{i=1}^N \xi_i  h(X_i) - \E \xi h \right| \leq \theta \|v-v_0\|_{L_2}^2.
$$
To that end, let $w_j=\sqrt{j}$, recall that $d_V=\sup_{v \in V} \|v\|_{L_2}$ and thus $d_{V_j}=s_j=r2^j$. Put
$$
u_j = \max\left\{8,\frac{\sqrt{N}\theta}{4c_4 \|\xi\|_{L_q}}\cdot \frac{2^jr}{\sqrt{j}} \cdot \frac{d_{V_j}}{\E\|G\|_{V_j}} \right\}
$$
and consider two cases: first, if $u_j > 8$ then clearly, $(*) \leq \theta s_j^2/4$ and
$$
Pr({\cal A}_j) \geq 1-c_5 \frac{\log^q N}{j^{q/2} N^{(q/2)-1}}-2\exp\left(-c_6(q,L)N  \frac{2^{2j}\theta^2}{j\|\xi\|_{L_q}^2} \cdot r^2\right).
$$
Alternatively, if $u_j=8$, then
$$
u_j^2 \left( \frac{\E\|G\|_{V_j}}{d_{V_j}}\right)^2 \geq c_7(q,L) r^2N\frac{2^{2j}\theta^2}{j\|\xi\|_{L_2}^2}.
$$
Also, by \eqref{eq:usefulcover-nostar}, $V_j$ has at most $|(V-v_0)\cap s_jD|$ extreme points. Since
\begin{align*}
\log |(V-v_0)\cap s_jD| & \leq  c_8\log(s_j/r) \log \cM\left(F_{v_0} \cap 4rD,(r/2)D\right)
\\
& \leq c_8\log(s_j/r) \eta_1 \sqrt{N} r,
\end{align*}
by standard properties of gaussian processes
\begin{align*}
\E\|G\|_{V_j} \leq & c_9d_{V_j} \cdot \log^{1/2}|(V-v_0) \cap s_jD|  \leq c_{10}s_j \log^{1/2}\left(\frac{2s_j}{r}\right) \eta_1 \sqrt{N}r
\\
= & c_{10}\eta_1  \sqrt{N} \sqrt{\frac{j}{2^j}} s_j^2.
\end{align*}
Hence, there are constants $c_{11}$ and $c_{12}$ that depend only on $q$ and $L$ for which
\begin{equation*}
\sup_{h \in V_j} \left|\frac{1}{N} \sum_{i=1}^N \xi_i h(X_i) - \E \xi h \right| \leq c_{11}\frac{\|\xi\|_{L_q}}{\sqrt{j}} \cdot \eta_1 \sqrt{\frac{j}{2^j}} s_j^2 \leq \theta s_j^2/4
\end{equation*}
if $\eta_1 \leq c_{12} \theta/\|\xi\|_{L_q}$.

Therefore, in both cases, there are constants $c_{13}$ and $c_{14}$ that depend only on $q$ and $L$, and with probability at least
$$
1-c_{13}\frac{\log^q N}{j^{q/2} N^{(q/2)-1}}-2\exp\left(-c_{14}N r^2 \eta_1^2 2^{j}\right),
$$
$$
\sup_{h \in V_j} \left|\frac{1}{N} \sum_{i=1}^N \xi_i h(X_i) - \E \xi h \right| \leq \theta s_j^2/4.
$$
The claim follows by applying the union bound to this estimate for $0 \leq j \leq j_0$.
\endproof

\vskip0.4cm

Next, let us turn to the infimum of the quadratic process
\begin{equation} \label{eq:inf-quad-process}
\inf_{\{v \in V : \|v-v_0\|_{L_2} \geq cr\}} \frac{1}{N}\sum_{i=1}^N \left(\frac{(v-v_0)}{\|v-v_0\|_{L_2}}\right)^2(X_i)
\end{equation}
where $r$ was selected in \eqref{eq:condition-on-r} for a well-chosen $\eta_2$ and where $c$ is a suitable constant.

\begin{Lemma} \label{thm:quadratic-local}
For every $L>1$ there exist constants $c_0,c_1$ and $c_2$ that depend only on $L$ for which the following holds. For every $v_0 \in V$, with probability at least $1-2\exp(-c_0N)$, if $v \in V$ and
$\|v-v_0\|_{L_2} \geq c_1r$ then
$$
\frac{1}{N}\sum_{i=1}^N (v-v_0)^2(X_i) \geq c_2\|v-v_0\|_{L_2}^2.
$$
\end{Lemma}

The proof of Lemma \ref{thm:quadratic-local} is similar to the one used in the analysis of the multiplier process: controlling relatively `large distances' in $F$, i.e., when $f \in F$ for which $\|f-v_0\|_{L_2} \geq (3/2)r_Q(\eta_2) \equiv s$; and then `small distances' in $V$, that is, $v \in V$ for which $r \leq \|v-v_0\|_{L_2} \leq s$ (again, one may assume that $r<r_Q(\eta_2)$).
\vskip0.3cm
For the constant $\eta_2$ (yet to be specified), one has
\begin{description}
\item{$\bullet$} $\E\|G\|_{(F-v_0) \cap s D} \leq \eta_2 N s $,
\item{$\bullet$} for every $2r<t<s$,
$$
\log \cN\left(({\rm star}(V-v_0)) \cap sD, tD\right) \leq c_0\log(2s/t) \cdot \eta_2^2 N,
$$
and
$$
\log |({\rm star}(V-v_0)) \cap sD | \leq c_0\log(2s/r) \cdot \eta_2^2 N.
$$
\end{description}

The required lower bound on the infimum of the quadratic process \eqref{eq:inf-quad-process} is based on estimates from \cite{Men-LWC} and \cite{Men-LWCG}, which will be formulated under the subgaussian assumption, rather than using the original (and much weaker) small-ball condition.
\begin{Theorem} \label{thm:quadratic-small-ball}
For every $L>1$ there are constants $\kappa_4$, $\kappa_5$ and $\kappa_6$ that depend only on $L$ for which the following holds.
Let $H$ be an $L$-subgaussian class that is star-shaped around zero. Set $H_\rho = H \cap \rho D$ and fix $\rho$ for which
$$
\E\|G\|_{H_\rho} \leq \kappa_4 \sqrt{N}\rho.
$$
Then, with probability at least $1-2\exp(-\kappa_5N)$,
$$
\inf_{\{h \in H : \|h\|_{L_2} \geq \rho\}} \frac{1}{N}\sum_{i=1}^N \left(\frac{h(X_i)}{\|h\|_{L_2}}\right)^2 \geq \kappa_6.
$$
\end{Theorem}

\vskip0.4cm

We will apply Theorem \ref{thm:quadratic-small-ball} to the class $H = (F-v_0) \cap s D$ (large distances) and then to $V_j={\rm star}\left((V-v_0) \cap s_jD\right)$
for $s_j=2^j r$ (small distances).

\begin{Lemma} \label{lemma:entropy-est}
There exist absolute constants $c_0$ and $c_1$ for which the following holds.
For every $s>\rho \geq c_0r$,
$$
\E\|G\|_{V_j \cap \rho D} \leq c_1\eta_2 \sqrt{N} \left(\rho \log^{3/2}(2s_j/\rho)+r\log^{3/2}(2s/r)\right).
$$
In particular, setting $\rho=s_j/2$ for $\eta_2=c_2\kappa_4$, one has
$$
\E\|G\|_{V_j \cap (s_j/2)D} \leq \kappa_4 \sqrt{N}(s_j/2).
$$
\end{Lemma}

\proof Fix $\rho < s_j$ and note that by Dudley's entropy integral bound (see, e.g., \cite{LT:91,vanderVaartWellner}),
\begin{align*}
& \E\|G\|_{V_j \cap \rho D} \leq c_1\int_0^\rho \log^{1/2} \cN \left(V_j \cap \rho D,tD\right)dt
\\
\leq & c_1\int_0^r \log^{1/2} \cN \left(V_j \cap \rho D,tD\right)dt + c_1\int_r^\rho \log^{1/2} \cN \left(V_j \cap \rho D,tD\right)dt.
\end{align*}
Applying \eqref{eq:useful-covering-star} and since
$$
V_j={\rm star}\left((V-v_0) \cap s_j D\right) \subset \left({\rm star}(V-v_0)\right) \cap s_j D,
$$
it follows that for $r<t<\rho$,
\begin{align*}
\log \cN (V_j \cap \rho D, rD) \leq & \log \cN \left( ({\rm star}(V-v_0))\cap \rho D,rD\right)
\\
\leq & c_2\log(2\rho/r) \cdot \sup_{x \in F}\log \cN\left((F-x) \cap 4rD,rD\right)
\\
\leq & c_2\log(2\rho/r) \cdot \eta_2^2 N.
\end{align*}
Moreover, by \eqref{eq:usefulcover-nostar},
$$
\log |(V-v_0) \cap s_j D| \leq c_2\log(2s_j/r) \cdot \eta_2^2 N =(*).
$$
Hence, $V_j$ is the union of at most $\exp(*)$ `intervals' of the from $[0,v-v_0]$, and
for $t \leq r$,
$$
\log \cN(V_j \cap \rho D, tD) \leq c_2\left(\eta_2^2 N \log(2s_j/r) + \log(2\rho/t)\right).
$$
Now the first part of the claim follows from integration, and the second part is an immediate outcome of the first.
\endproof

\noindent{\bf Proof of Lemma \ref{thm:quadratic-local}.}
Combining Theorem \ref{thm:quadratic-small-ball} and Lemma \ref{lemma:entropy-est} for $\eta_2=c_0 \kappa_4$, it follows that with probability at least $1-2\exp(-\kappa_5 N)$, if $v \in V$ and $s_j/2 \leq \|v-v_0\|_{L_2} \leq s_j$,
\begin{equation} \label{eq:inf-est-in-proof}
\frac{1}{N}\sum_{i=1}^N (v-v_0)^2(X_i) \geq \kappa_6 \|v-v_0\|_{L_2}^2.
\end{equation}
Repeating this argument for $s_j =2^j r$ and then applying it to the set $F_{v_0} \cap sD$ for $s=(3/2)r_Q(\eta_2)$, it follows that if $\log_2(s/r) \leq \exp(\kappa_5 N/2)$ then with probability at least $1-2\exp(-\kappa_5 N/2)$, \eqref{eq:inf-est-in-proof} holds for every $v \in V$ that satisfies $\|v-v_0\|_{L_2} \geq c_1r$.
\endproof

\vskip0.4cm

With all the ingredients in place, we may now conclude the proof of the upper estimate.

Fix $f_0 \in F$ and set $Y=f_0(X)+W$ for $W \in L_q$ that is orthogonal to ${\rm span}(F)$. Let $r$, $V$ and $v_0$ as above. Clearly, for every $v \in V$,
\begin{equation} \label{eq:excess-for-V}
\|v-Y\|_{L_2}^2 = \|W\|_{L_2}^2 + \|v-f_0\|_{L_2}^2,
\end{equation}
and thus $\|v_0-f_0\|_{L_2} \leq r$. Moreover, for every $v \in V$, $\E W \cdot (v-v_0)(X)=0$ and
\begin{align*}
\left|\E(v_0(X)-Y)(v-v_0)(X)\right| = & \left|\E(v_0-f_0)(X) \cdot (v-v_0)(X)\right|
\\
\leq & \|v_0-f_0\|_{L_2} \cdot \|v-v_0\|_{L_2} \leq r \|v-v_0\|_{L_2}.
\end{align*}

By Lemma \ref{thm:quadratic-local}, with probability at least $1-2\exp(-\kappa_5N/2)$, if $v \in V$ and $\|v-v_0\|_{L_2} \geq c(L)r$, then
$$
\frac{1}{N}\sum_{i=1}^N (v-v_0)^2(X_i) \geq \kappa_6 \|v-v_0\|_{L_2}^2.
$$
Using the notation of Lemma \ref{lemma:local-multiplier}, set $\theta=\kappa_6/4$ and $\eta_1=c_0(q,L)\theta/\|W\|_{L_q}$. Hence, there are constants $c_1$ and $c_2$ that depend only on $q$ and $L$, for which, with probability at least
$$
1-c_1\frac{\log^q N}{N^{((q/2)-1)}} - 2\exp(-c_2\eta_1^2r^2N),
$$
for every $v \in V$, $\|v-v_0\|_{L_2} \geq 2r$,
$$
\left|\frac{1}{N}\sum_{i=1}^N (v_0(X_i)-Y_i)(v-v_0)(X_i)-\E(v_0(X)-Y)(v-v_0)(X)\right| \leq \frac{\kappa_6}{4} \|v-v_0\|_{L_2}^2.
$$
On the intersection of the two events and for a constant $c_3=c_3(q,L)$, if $\|v-v_0\|_{L_2} \geq c_3r$ then
\begin{align*}
P_N {\cal L}_v^V = & \frac{1}{N}\sum_{i=1}^N (v-v_0)^2(X_i) + \frac{2}{N} \sum_{i=1}^N (v_0(X_i)-Y_i)(v-v_0)(X_i)
\\
\geq & \frac{1}{N}\sum_{i=1}^N (v-v_0)^2(X_i) - 2|\E (v_0(X)-Y)(v-v_0)(X)|
\\
& - 2\left|\frac{1}{N} \sum_{i=1}^N (v_0(X_i)-Y_i)(v-v_0)(X_i) - \E (v_0(X)-Y)(v-v_0)(X)\right|
\\
\geq & \kappa_6 \|v-v_0\|_{L_2}^2 - 2r\|v-v_0\|_{L_2} - (\kappa_6/4) \|v-v_0\|_{L_2}^2 \geq (\kappa_6/4)\|v-v_0\|_{L_2}^2.
\end{align*}
Thus, for every such sample, the empirical minimizer $\hat{v} \in V$ satisfies that
$$
\|\hat{v}-v_0\|_{L_2} \leq c_4r.
$$
And, since $W$ is orthogonal to ${\rm span}(F)$,
\begin{align*}
& \E \left({\cal L}_{\hat{v}}^F |(X_i,Y_i)_{i=1}^N\right) =\|\hat{v}-Y\|_{L_2}^2-\|f_0-Y\|_{L_2}^2 = \|\hat{v}-f_0-W\|_{L_2}^2 - \|W\|_{L_2}^2
\\
= & \|\hat{v}-f_0\|_{L_2}^2 -2\E W \cdot (\hat{v}-f_0)(X) \leq \left(\|\hat{v}-v_0\|_{L_2} + \|v_0-f_0\|_{L_2}\right)^2
\leq (1+c_4)^2 r^2.
\end{align*}
\endproof

\section{The lower bound} \label{sec:lower}
The lower estimates presented below are based on a volumetric argument. The idea is that if a learning procedure is `too successful', a well-separated subset of $F$ endows a well-separated subset in $\R^N$ (a set that depends on $X_1,...,X_N$). However, because of some volumetric constraint, there is not `enough room' for such a separated set to exist, leading to a contradiction.

The notions of volume are different in the two estimates: one is based on the Lebesgue measure while the other is determined by the choice of the `noise' $W$, which is, in our case, gaussian.

\begin{Definition}
Let $F$ be a class of functions and assume that $\mathbb{X}=(x_1,...,x_N) \in \Omega^N$. For every $f \in F$, set
$$
{\cal K}(f,\mathbb{X}) = \{h \in F : h(x_i)=f(x_i) \ {\rm for \ every \ } 1 \leq i \leq N\}.
$$
The set ${\cal K}(f,\mathbb{X})$ is called the {\it version space} of $F$ associated with $f$ and $\mathbb{X}$.
\end{Definition}

In other words, ${\cal K}(f,\mathbb{X})$ consists of all the functions in $F$ that agree with $f$ on $\mathbb{X}$. Naturally, in the context of learning, $\mathbb{X}$ is a random sample $(X_i)_{i=1}^N$, selected according to the underlying measure $\mu$.

The diameter of the version space is a reasonable choice for a lower bound on the performance of any learning procedure: if $Y_i=f(X_i)+W_i$, a learning procedure cannot distinguish between $f$ and any other function in the version space associated with $f$ and $(X_i)_{i=1}^N$. Hence, the largest typical diameter of a version space should be a lower estimate on the performance of any learning procedure, as the following well-known fact shows (see, e.g., \cite{LM13}).

\begin{Theorem} \label{thm:lower-noise-free}
Given a random variable $W$, for every $f \in F$ set $Y^f=f(X)+W$. If $\Psi$ is a learning procedure, then
\begin{equation*}
  \sup_{f \in F}Pr\left(\|\Psi((Y^f_i,X_i)_{i=1}^N)-f\|_{L_2(\mu)} \geq  \frac{1}{4}{\cal K}(f,{\mathbb X}) \right) \geq 1/2,
\end{equation*}
where the probability is relative to the product measure endowed on $(\Omega \times \R)^N$ by the $N$-product of the joint distribution of $X$ and $W$.
\end{Theorem}

Clearly, if $W$ is orthogonal to ${\rm span}(F)$, then for every $h \in F$ and every target $Y^f$, $\E {\cal L}_h^f=\|h-f\|_{L_2}^2$. Thus, the largest typical diameter of a version space ${\cal K}(f,{\mathbb X})$ is a lower bound on ${\cal E}_p$ for the set of admissible targets ${\cal Y}=\{f(X)+W : f \in F\}$.

This leads to the following question:
\begin{Question} \label{qu:lower-est-version-space}
Given a class $F$ defined on a probability space $(\Omega,\mu)$, $f \in F$ and $\mathbb{X}=(x_1,...,x_N) \subset \Omega^N$, find a lower estimate on
$$
{\rm diam}\left({\cal K}(f,\mathbb{X}),L_2(\mu)\right).
$$
\end{Question}

One situation in which Question \ref{qu:lower-est-version-space} is of independent interest is when $T \subset \R^n$ is a convex body (i.e., a convex, centrally-symmetric set with a nonempty interior) and $F=\left\{\inr{t,\cdot} : t \in T\right\}$ is the class of linear functionals associated with $T$.
For every $x_1,...,x_N \in \R^n$ set $\mathbb{X}=(x_1,....,x_N)$, and let $\Gamma_{\mathbb X}=\sum_{i=1}^N \inr{x_i,\cdot}e_i$ be the matrix whose rows are $x_1,...,x_N$. Thus,
$$
{\cal K}(0,\mathbb{X})={\rm ker}(\Gamma_{\mathbb X}) \cap T.
$$
If $\mu$ is an isotropic, $L$-subgaussian measure on $\R^n$, one may show that with probability at least $1-2\exp(-c_0N)$,
\begin{equation} \label{eq:PT}
{\rm diam}({\cal K}(0,\mathbb{X}),L_2(\mu)) \leq 2r_Q(c_1(L))
\end{equation}
(see \cite{MR2373017}). This extends the celebrated result of Pajor and Tomczak-Jaegermann \cite{MR941809,MR845980}, that \eqref{eq:PT} holds for the Haar measure on $S^{n-1}$ (and thus, also for the gaussian measure on $\R^n$).
\vskip0.3cm
It turns out that \eqref{eq:PT} is not far from optimal:
\begin{Theorem} \label{thm:kernel}
There exists an absolute constant $c$ for which the following holds.
Let $F \subset L_2(\mu)$ be a convex and centrally-symmetric set. If
$$
\log {\cal M}(F \cap 2rD,(r/4)D) \geq cN,
$$
then for every $\mathbb{X}=(x_1,...,x_N)$,
$$
{\rm diam}\left({\cal K}(0,\mathbb{X}),L_2(\mu)\right) \geq r/8.
$$
\end{Theorem}
\vskip0.3cm
Since $F$ is convex and centrally-symmetric, $F-F =2F$ and $0 \in F$. Therefore,
\begin{align*}
& \cM(F \cap 4rD,(r/2)D) \leq \sup_{x \in F} \cM\left((F-x) \cap 4rD,(r/2)D\right)
\\
\leq & \cM\left((F-F) \cap 4rD,(r/2)D \right) = \cM\left( F \cap 2rD,(r/4)D \right).
\end{align*}
Hence, Theorem \ref{thm:kernel} shows that if $\gamma_Q(c,0) > r$ then for every $\mathbb{X}=(x_1,...,x_N)$, ${\rm diam}({\cal K}(0,\mathbb{X}),L_2(\mu)) \geq r/8$.
In particular, for every $W \in L_2$ that is orthogonal to ${\rm span}(F)$, the best possible error rate in $F$ that holds for every target $Y^f=f(X)+W$, is at least $\gamma_Q^2(c,0) \geq c_1 \gamma_Q^2(c)$.

\vskip0.3cm

\proof
Let $f_1,...,f_m$ be $r/4$-separated in $F \cap 2rD$. Set
$$
A_i=\frac{f_i}{2}+\frac{1}{32}(F \cap 2rD),
$$
and observe that $A_i \subset F \cap 2rD$. Also, for every $h \in A_i$, $\|(f_i/2)-h\|_{L_2} \leq r/16$; therefore, if $h_i \in A_i$ and $h_\ell \in A_\ell$, then $\|h_i-h_\ell\|_{L_2} \geq r/8$.

Fix $\mathbb{X}=(x_1,...,x_N)$ and for $A \subset F$ set
$$
P_{\mathbb{X}}(A)=\left\{ (h(X_i))_{i=1}^N : h \in A\right\} \subset \R^N,
$$
the coordinate projection of $A$ associated with $\mathbb{X}$. Clearly, for every $1 \leq i \leq m$,
\begin{equation} \label{eq:separated-empirical}
P_{\mathbb{X}}(A_i)=\frac{1}{2}(f_i(x_j))_{j=1}^N + \frac{1}{32} P_{\mathbb{X}}(F \cap 2rD).
\end{equation}

Consider two possibilities. First, if there are $i \not= \ell$ for which $P_{\mathbb{X}}(A_i) \cap P_{\mathbb{X}}(A_\ell) \not= \emptyset$, there are $h_i \in A_i$ and $h_\ell \in A_\ell$ that satisfy $h_i-h_\ell \in {\cal K}(0,\mathbb{X})$, thus showing that ${\rm diam}({\cal K}(0,\mathbb{X}),L_2(\mu)) \geq r/8$.

Otherwise, the sets $P_{\mathbb{X}}(A_i)$ are disjoint subsets of $P_{\mathbb{X}}(F \cap 2rD)$. And, setting $T=P_{\mathbb{X}}(F \cap 2rD)$, \eqref{eq:separated-empirical} implies that $\cM(T,T/32) \geq m$. Since $T$ is a convex, centrally symmetric subset of $\R^N$, a standard volumetric argument shows that $\cM(T,T/32) \leq \exp(cN)$ for a suitable absolute constant $c$.
Thus, if $m > \exp(cN)$, ${\rm diam}\left({\cal K}(0,\mathbb{X}),L_2\right)\geq r/8$, as claimed.
\endproof

\vskip0.4cm

The final result of this section is the `noise-dependent' lower bound.

\begin{Theorem} \label{thm:lower-bound}
There exist absolute constants $c_1$ and $c_2$ for which the following holds. Let $F \subset L_2(\mu)$ be a convex, centrally-symmetric class of functions, set $W$ to be a centred normal random variable that is independent of $X$, and for every $f \in F$, put $Y^f=f(X)+W$. If $\Psi$ is a learning procedure that performs with confidence of at least $7/8$ for every $Y^f$, there is some $Y^f$ for which
$$
{\cal E}_p \geq c_1 \gamma_M^2\left(\frac{c_2}{\|W\|_{L_2}}\right).
$$
\end{Theorem}

Stronger versions of Theorem \ref{thm:lower-bound} (without the assumption that $F$ is convex and centrally-symmetric) may be proved in several different ways: using information theoretic tools  (see, Theorem 2.5 in \cite{MR2724359}), or, alternatively, by applying the gaussian isoperimetric inequality as in \cite{LM13}. Both these arguments are rather restrictive, because they relay on rather special properties of the noise.

Although the proof we present below is also for a gaussian noise, the argument is less restrictive and may be extended to other choices of noise (e.g. when $W$ is log-concave rather than gaussian). The argument is essentially the same as Talagrand's proof of the dual-Sudakov inequality \cite{LT:91}, and as such is volumetric in nature: obtaining a lower bound on the measure of a shift of a centrally-symmetric set in terms of the Euclidean norm of the shift.

\begin{Lemma} \label{lemma:lemma-from-dual-sudakov}
Let $A \subset \R^N$ be centrally symmetric and set $z \in \R^N$. If $\nu$ is the centred gaussian measure on $\R^N$ with covariance $\sigma^2 I_N$ and $| \ |$ denotes the Euclidean norm on $\R^n$, then
$$
\nu(z+A) \geq \exp\left(-\frac{|z|^2}{2\sigma^2}\right) \nu (A).
$$
\end{Lemma}
\proof
A change of variables shows that
\begin{align*}
\nu(z+A) = & \frac{1}{(2\pi \sigma)^{N/2}} \int_{z+A} \exp\left(-\frac{|x|^2}{2\sigma^2}\right)dx
= \frac{1}{(2\pi \sigma)^{N/2}} \int_{A}\exp\left(-\frac{|t+z|^2}{2\sigma^2}\right) dt
\\
= & \exp\left(-\frac{|z|^2}{2\sigma^2}\right) \cdot \frac{1}{(2\pi \sigma)^{N/2}} \int_{A} \exp\left(\frac{\inr{z,t}}{\sigma^2}\right) \cdot \exp\left(-\frac{|t|^2}{2\sigma^2}\right) dt =(*).
\end{align*}
Let $\E_{\nu|A}$ be the expectation with respect to the gaussian measure $\nu$, conditioned on $A$.  Thus,
$$
(*)=\exp\left(-\frac{|z|^2}{2\sigma^2}\right) \nu(A) \cdot \E_{\nu|A}\exp\left(-\frac{\inr{z,t}}{\sigma^2}\right).
$$
Since $A$ is symmetric, $\E_{\nu|A}\inr{z,t}=0$, and by Jensen's inequality
\begin{equation*}
(*) \geq \exp\left(-\frac{|z|^2}{2\sigma^2}\right) \nu(A).
\end{equation*}
\endproof

\noindent{\bf Proof of Theorem \ref{thm:lower-bound}.} Let $\Psi$ be a learning procedure that performs with accuracy ${\cal E}_p$ for every target $Y^f = f(X)+W$ for $f \in F$ and $W \sim {\cal N}(0,\sigma^2)$ that is independent of $X$. Note that for the target $Y^f$, the true minimizer in $F$ is $f^*=f$ and for every $h \in F$,
$$
\E {\cal L}_h = \E (h(X)-Y^f)^2 - \E(f^*(X)-Y^f)^2 = \|h-f\|_{L_2}^2.
$$
Thus, if $\tau=(x_i,y_i)_{i=1}^N \in (\Omega \times \R)^N$ is a sample on which $\Psi$ performs with accuracy ${\cal E}_p$ relative to the target $Y^f$, then $\|\Psi(\tau)-f^*\|_{L_2}^2 \leq {\cal E}_p$.

Let $(f_j)_{j=1}^m$ be a subset of $F \cap 4rD$ that is $r/2$ separated in $L_2(\mu)$ for $(r/2)^2 =9 {\cal E}_p$ and fix $\mathbb{X}=(x_1,...,x_N) \in \Omega^N$.

For every $1 \leq j \leq m$, put
$$
A_j(\mathbb{X}) = \left\{ (w_i)_{i=1}^N : \Psi\left((x_i,f_j(x_i)+w_i)_{i=1}^N\right) \in f_j+\sqrt{{\cal E}_p}D \right\} \subset \R^N,
$$
i.e., $A_j(\mathbb{X})$ consists of all the vectors $(w_i)_{i=1}^N  \in \R^N$, for which, upon receiving the data $(x_i,f_j(x_i)+w_i)_{i=1}^N$, $\Psi$ selects a point whose $L_2$ distance to $f_j$ is at most $r/6=\sqrt{{\cal E}_p}$.

Let $\nu$ be the centred gaussian measure on $\R^N$ with covariance $\sigma^2 I_N$. Since $W$ is a centred gaussian random variable with variance $\sigma^2$, $(w_i)_{i=1}^N$ is distributed according to $\nu$, and since it is independent of $X$,
if $\Psi$ performs with accuracy ${\cal E}_p$ and with probability at least $7/8$, it is evident that
\begin{align*}
& \mu^N \otimes \nu \left(\left\{ (x_i,w_i)_{i=1}^N : \Psi((x_i, f_j(x_i)+w_i)_{i=1}^N) \in f_j + \sqrt{{\cal E}_p} D\right\}\right)
\\
= & \mu^N \otimes \nu \left(\left\{ (x_i,w_i)_{i=1}^N : (w_i)_{i=1}^N \in A_j(\mathbb{X})\right\} \right) \geq 7/8.
\end{align*}
A standard Fubini argument shows that there is an event ${\cal C}_j \subset \Omega^N$ of $\mu^N$ probability at least $1/2$, and for every $\mathbb{X}=(x_i)_{i=1}^N \in {\cal C}_j$, $\nu\left(A_j(\mathbb{X})\right) \geq 3/4$. Observe that if $\mathbb{X} \in {\cal C}_j$ then by the symmetry of $\nu$, $\nu\left(-A_j(\mathbb{X})\right) \geq 3/4$, and the centrally-symmetric set $A_j(\mathbb{X}) \cap - A_j(\mathbb{X}) \subset A_j(\mathbb{X})$ satisfies that
$$
\nu\left(A_j(\mathbb{X}) \cap -A_j(\mathbb{X})\right) \geq 1/2.
$$

Let $z_j=(f_j(x_i))_{i=1}^N$. If $\mathbb{X} \in {\cal C}_j \cap {\cal C}_\ell$,
the sets $z_j+A_j(\mathbb{X})$ and $z_\ell+A_\ell(\mathbb{X})$ are disjoint, because $\Psi$ maps $z_j+A_j(\mathbb{X})$ to an $r/6$-neighbourhood of $f_j$ and $z_\ell+A_\ell(\mathbb{X})$ to an $r/6$-neighbourhood of $f_\ell$ -- but $\|f_j-f_\ell\|_{L_2} \geq r/2$.
Therefore
$$
\sum_{j=1}^m \IND_{{\cal C}_j}(\mathbb{X}) \nu\left(z_j+\left(A_j(\mathbb{X}) \cap -A_j(\mathbb{X})\right) \right) \leq 1;
$$
integrating with respect to $\mu^N$,
$$
\sum_{i=1}^m \E_{\mathbb{X}} \IND_{{\cal C}_j}(\mathbb{X}) \nu\left(z_j+\left(A_j(\mathbb{X}) \cap -A_j(\mathbb{X})\right) \right) \leq 1,
$$
and all that remains is to control $\E_{\mathbb{X}} \IND_{{\cal C}_j}(\mathbb{X}) \nu\left(z_j+\left(A_j(\mathbb{X}) \cap -A_j(\mathbb{X})\right) \right)$ from below.

Applying Lemma \ref{lemma:lemma-from-dual-sudakov},
\begin{align*}
\nu\left(z_j+\left(A_j(\mathbb{X}) \cap -A_j(\mathbb{X})\right) \right) \geq & \exp\left(-\frac{|z_j|^2}{2\sigma^2}\right) \cdot \nu\left(A_j(\mathbb{X}) \cap -A_j(\mathbb{X})\right)
\\
= & \exp\left(-\frac{1}{2\sigma^2}\sum_{i=1}^N f_j^2(x_i)\right) \cdot \nu\left(A_j(\mathbb{X}) \cap -A_j(\mathbb{X})\right).
\end{align*}
By Chebychev's inequality and recalling that $\|f_j\|_{L_2} \leq 4r$,
$$
\mu^N \left\{ \sum_{i=1}^N f_j^2(X_i) \leq c_0Nr^2\right\} \geq 3/4
$$
for an appropriate choice of an absolute constant $c_0$ and for every $1 \leq j \leq m$. Thus, on an event of $\mu^N$ measure at least $1/4$, $\mathbb{X} \in {\cal C}_j$, $\nu\left(A_j(\mathbb{X}) \cap -A_j(\mathbb{X})\right) \geq 3/4$ and $\sum_{i=1}^N f_j^2(X_i) \leq c_0Nr^2$; therefore,
\begin{equation*}
\E_{\mathbb{X}} \IND_{{\cal C}_j}(\mathbb{X})\nu\left(z_j+\left(A_j(\mathbb{X}) \cap -A_j(\mathbb{X}) \right)\right) \geq c_1 \exp\left(-c_0 \frac{N r^2}{2\sigma^2}\right).
\end{equation*}
Hence, $\log m \leq c_2^2 Nr^2/\sigma^2$, i.e., $\log \cM(F \cap 4rD,(r/2)D) \leq (c_2/\sigma)^2Nr^2$, implying that ${\cal E}_p \geq c_3\gamma_M^2(c_2/\sigma)$.
\endproof

\section{Some Remarks} \label{sec:remarks}
We begin this section with an example of `natural' sets, for which there is a true gap between the two sets of parameters: $r_Q/r_M$ and $\gamma_Q/\gamma_M$.

Let $T \subset \R^n$ be a convex body in $\R^n$ (i.e., a convex, centrally-symmetric set with a nonempty interior), put $F=\{\inr{t,\cdot} : t \in T\}$, the class of linear functionals associated with $T$ and set $\mu$ to be the gaussian measure on $\R^n$.

It is straightforward to verify that for every $r>0$, $(F \cap rD,L_2(\mu))$ is isometric to $(T \cap rB_2^n,\ell_2^n)$, where $B_2^n$ is the Euclidean unit ball in $\R^n$. Let $1 \leq p <2$, and set $T=B_p^n$, the unit ball in $\ell_p^n=(\R^n,\| \ \|_{\ell_p})$. One may show (see \cite{LM13}) that when $p=1$, $r_M$ and $\gamma_M$ are equivalent, as are $r_Q$ and $\gamma_Q$. However, such an equivalence is no longer true for $1<p<2$ (of course, as long as $p>1+1/\log n$ -- otherwise, $\ell_p^n$ is equivalent to $\ell_1^n$).

To see how that gap between the `global' and `local' parameters is exhibited in $B_p^n$ for $1<p<2$, let $x=(x_i)_{i=1}^n \in B_p^n$ and set $(x_i^*)_{i=1}^n$ to be the non-increasing rearrangement of $(|x_i|)_{i=1}^n$; thus, $x_i^* \leq i^{-1/p}$. Recall the well known fact (see, e.g., \cite{MR2371614}), that $\E\|G\|_{B_p^n \cap rB_2^n}$ is equivalent to
\begin{equation*}
c_1(p)
\begin{cases}
n^{1-1/p}  & \mbox{if} \ \ r \geq c_2(p)n^{-(1/p-1/2)},
\\
rn^{1/2} & \mbox{if} \ \ r \leq c_2(p)n^{-(1/p-1/2)}.
\end{cases}
\end{equation*}
Thus, if $N \leq n^{2/p}$,
$$
r_M \sim n^{1/2-1/2p}/N^{1/4} \geq c_2(p)n^{-(1/p-1/2)}.
$$

Let us consider the case in which $1 > r \gg c_2(p)n^{-(1/p-1/2)}$. Set $\ell=(1/r)^{2p/(2-p)}$ and observe that
$$
B_p^n \cap 4rB_2^n \subset \left\{x \in \R^n \ : \ x_i^* \leq 4r/i^{1/2} \ \ {\rm if} \ i \leq \ell, \ \ {\rm and} \ \ x_i^* \leq 1/i^{1/p} \ \ {\rm if} \ i > \ell\right\}.
$$
Clearly, for a well-chosen constant $c_3$ one has $\sum_{i \geq c_3\ell} i^{-2/p} \leq r^2/100$, and
$$
B_p^n \cap 4rB_2^n \subset \bigcup_{|I| = c_3\ell} \left(c_4 rB_{2,\infty}^I + B_{p,\infty}^{I^c}\right),
$$
where $B_{q,\infty}^I$ is the unit ball in $\R^I$ endowed with the weak $\ell_{q,\infty}$ norm\footnote{Recall that for $x \in \R^n$, $\|x\|_{q,\infty} \leq A$ if and only if, $\sup_{i \geq 1} i^{1/q}x_i^* \leq A$.}. In particular, if $|I| = c_3\ell$,  $B_{p,\infty}^{I^c} \subset (r/10)B_2^{I^c}$ and the impact of those `small' coordinates on Euclidean distances is negligible:
$$
\cM\left(B_p^n \cap 4r B_2^n, rB_2^n\right) \leq \cM\left(\bigcup_{|I|= c_3\ell} c_4rB_2^I,(r/2)B_2^n\right) \leq \binom{n}{c_3\ell}c_4^{c_3\ell}.
$$

Hence, separation at scale $r$ occurs only because of the largest $\sim \ell$ coordinates of the vectors involved.

On the other hand, the contribution of those `large' coordinates to $\E\|G\|_{B_p^n \cap 4rB_2^n}$ is equally negligible. Indeed, if $T = \bigcup_{|I|=m} \alpha r B_2^I$ for some $m \leq n/2$ and $\alpha \geq 1$, it is standard to verify that
$$
\E \sup_{t \in T} \sum_{i=1}^n g_i t_i \leq c_5 \alpha r m^{1/2} \cdot \log^{1/2}\left(\frac{en}{m}\right) =(*).
$$
If $m=c_3\ell=c_3(1/r)^{2p/(2-p)}$ and since $r \gg c_2(p)n^{-(1/p-1/2)}$, it follows that
$$
(*) \ll n^{1-1/p} \sim \E\|G\|_{B_p^n \cap rB_2^n}.
$$
Thus, as long as $r$ is significantly larger than $c_2(p)n^{-(1/p-1/2)}$, the gaussian average of the intersection body $B_p^n \cap 4rB_2^n$ originates from the `small coordinates' in the monotone rearrangement, and in particular, from vectors whose Euclidean norm is significantly smaller than $r$. Such vectors are `invisible' to $\gamma_M$, which is why $\gamma_M$ is much smaller than $r_M$.

\vskip0.5cm
\subsection{The role of fixed points}

Fixed points are encountered frequently in Empirical Processes and Statistics literature, and almost always with the same goal: obtaining `relative' upper bounds on various empirical processes. To obtain such bounds, one has to compare the oscillation (i.e., the behaviour of the process indexed by $(F-F) \cap rD$) with some function of $r$.

One usually obtains upper bounds on the oscillation via a symmetrization argument, leading to a sample-dependent Bernoulli process. Thus, the standard outcome is a fixed point equation, linking an entropy integral relative to the random $L_2$ metric and generated by the sample $X_1,...,X_N$, with the desired function of $r$ (see \cite{vanderVaartWellner} for numerous examples).

Still within the realm of entropy integrals, it is possible to impose additional structure on the problem, which allows one to replace the empirical $L_2$ (random) metrics with the global $L_2(\mu)$ metric. For example, a fixed point equation with the same normalization as $r_M$ may be found in \cite{MR1240719}, where the setup allows the transition between the random metric and the deterministic one -- but the `philosophy' of the proof is the same: it is based on an entropy integral.
 
Since the entropy integral is only upper estimate on the supremum of the empirical process in question -- regardless of the underlying assumptions, it is often loose. Therefore, one would like to find a general argument bypassing the whole mechanism of entropy integrals.

As a first step, and because it is natural to expect that the empirical processes in question converges to a gaussian limit, one may try a `gaussian'-based fixed point, which relies on $\E\|G\|_{(F-F) \cap rD}$, rather than on an entropy integral bound. And, indeed, under a subgaussian assumption, the results of \cite{LM13} lead to the gaussian-based $r_M$ and $r_Q$.

Our results show that $r_M$ and $r_Q$ are not the end of the story and can be improved -- at least for the special learning problems we consider. The `right' fixed points should involve the smaller local entropy estimates rather than the oscillation of the gaussian process.
\vskip0.3cm
One fixed point that seems closer in nature to $\gamma_M$ than to $r_M$ may be found in the celebrated work of Yang and Barron \cite{MR1742500}, though a closer inspection shows that this impression is inaccurate.

Comparing \cite{MR1742500} to our results is somewhat unnatural because the setup in \cite{MR1742500} is completely different: a function class consisting of uniformly bounded functions and an independent gaussian noise, both of which are crucial to the proof (see Section 3.2 in \cite{MR1742500}). Also, the upper estimate is an existence result of a `good' procedure -- rather than a specific choice of a procedure; the estimate holds in expectation and not with high probability; and it does not tend to zero with the `noise level' of the problem.

All these differences are significant, but are still not a conclusive indication that the nature of the complexity parameter in \cite{MR1742500} is different from ours. That indication is the key to the results in \cite{MR1742500}: the assumption that the underlying class `large' -- in the sense that
\begin{equation} \label{eq:Yang-Barron}
\liminf_{\eps \to 0} \frac{\log {\cal M}(F,(\eps/2) D)}{\log {\cal M}(F,\eps D)} >1.
\end{equation}
One should note that this assumption immediately excludes all the modern high-dimensional problems, involving classes indexed by subsets of $\R^n$. Indeed, for any convex subset of $\R^n$, the liminf above is $1$ rather than strictly greater than $1$.
\vskip0.3cm
Equation \eqref{eq:Yang-Barron} has two significant implications:
\begin{description}
\item{$\bullet$} The $r/2$ log-covering numbers of $F$ and of $F \cap rD$ are equivalent, which means that one may replace the local sets $F \cap rD$ with $F$ in the definition of the fixed points. This makes the proof of the upper bound simpler.
\item{$\bullet$} It essentially restricts the setup to classes that have polynomial entropy, which is a considerably narrower scenario. Indeed, for the sake of brevity let us ignore cases in which
$$
\limsup_{\eps \to 0} \frac{\log {\cal M}(F,(\eps/2) D)}{\log {\cal M}(F,\eps D)} = L \geq 4
$$
(if $L > 4$ then the gaussian process $\{G_f : f \in F\}$ is not bounded and the class $F$ is not subgaussian, while if $L=4$ an entropy estimate is not enough to determine whether the gaussian process is bounded and thus requires a more subtle analysis). When $L <4$ and because the entropy of the `local' set $F \cap rD$ is equivalent to the entropy of $F$, it follows that there are $0<q_1 \leq q_2 <2$ for which, for every $\eps \leq R$ small enough,
$$
\left(\frac{R}{\eps}\right)^{q_1} \lesssim \frac{\log {\cal M}(F \cap RD,(\eps/2) D)}{ \log {\cal M}(F \cap RD,\eps D)} \lesssim \left(\frac{R}{\eps}\right)^{q_2}.
$$
Using Dudley's entropy integral for the upper bound and Sudakov's minoration for the lower one, it is straightforward to verify that
$$
\E\|G\|_{F \cap RD} \sim_{q_1,q_2}  R \log^{1/2} {\cal M}(F \cap RD, (R/2) D).
$$
Therefore, the `global' parameters $r_M$ and $r_Q$ are equivalent to the local ones $\gamma_M$ and $\gamma_Q$; in fact, the `local' and `global' parameters are even equivalent to the ones defined via the entropy integral. Thus, the typical situation in \cite{MR1742500} is very different from the problems studied here -- mainly because of \eqref{eq:Yang-Barron}.
\end{description}

\begin{footnotesize}

\bibliographystyle{plain}

\bibliography{biblio}
\end{footnotesize}

\end{document}